# Texture feature extraction in the spatial-frequency domain for content-based image retrieval

Nadia Baaziz, Omar Abahmane and Rokia Missaoui

*Department of Computer Science and Engineering*

*Université du Québec en Outaouais*

*101 rue Saint Jean Bosco, C.P. 1250,*

*Gatineau, Québec, J8X 3X7 Canada*

Emails :{ nadia.baaziz@uqo.ca; abao01@uqo.ca; rokia.missaoui@uqo.ca}

**Abstract.** The advent of large scale multimedia databases has led to great challenges in content-based image retrieval (CBIR). Even though CBIR is considered an emerging field of research, however it constitutes a strong background for new methodologies and systems implementations. Therefore, many research contributions are focusing on techniques enabling higher image retrieval accuracy while preserving low level of computational complexity. Image retrieval based on texture features is receiving special attention because of the omnipresence of this visual feature in most real-world images. This paper highlights the state-of-the-art and current progress relevant to texture-based image retrieval and spatial-frequency image representations. In particular, it gives an overview of statistical methodologies and techniques employed for texture feature extraction using most popular spatial-frequency image transforms, namely discrete wavelets, Gabor wavelets, dual-tree complex wavelet and contourlets. Indications are also given about used similarity measurement functions and most important achieved results.

## 1. Introduction

Due to the widespread use of digital technologies and Internet networks, the activity of producing, retrieving and distributing multimedia data becomes a frequent but still challenging task of retrieving data from large scale multimedia databases with satisfactory accuracy and performance rates. Digital image databases are growing very fast and cover a wide variety of application areas, such as aerial, geography and medical imaging, security and personal identification, quality inspection, remote sensing, cultural heritage management and preservation and so on. In this context, text-based image retrieval using keyword indexes has rapidly shown its limitations but has paved the way for more promising alternatives such as content-based image retrieval (CBIR). Using various low-level visual features (color, shape, texture,…) to index images, CBIR has been a very active research field during the last decade and has gained a lot of attention from researchers worldwide [1, 2].

Designing online CBIR systems proves to be a challenging task since key objectives rely on retrieval techniques that achieve a compromise between three main performance factors: the accuracy of retrieval results based on relevant visual features and similarity measures, the retrieval time that has to be optimized to offer acceptable user wait time and the processing complexity that usually impacts on computation and resources. A typical CBIR system consists of two main tasks, namely feature extraction and similarity measurement. During the feature extraction step, each target image is processed and a set of feature vectors describing its content are computed to form the database visual index. Retrieving similar images to the user query image is done through the calculation of predefined similarity measures (distance metrics) between the query image feature vectors and the database visual index features. Images that are mostly similar to the query image are then retrieved. It is important to mention that the success of CBIR is strongly related on the choice of an efficient similarity metric and the development of feature extraction methods that achieve powerful characteristic discrimination while providing feature representation with reduced dimensionality. Commonly used low level visual features for image indexing include color, shape, texture or any combination of them. Following the limitations related to the use of global characteristics, such as color, and due to the fact that textures are omnipresent in most real-world images, texture features have gained more significance in CBIR. Therefore, the research presented in this paper explores some of the recent techniques, appeared in the last ten years, which exploit spatial-frequency image transforms and statistical modeling for texture feature analysis and extraction in the context of CBIR applications. This research is accomplished as a part of an ongoing project aiming at extending our CBIR prototype MIRA (Multimedia Information Retrieval Application) by integrating texture-based image retrieval [3]. The remainder of this paper is organized as follows. Section 2 describes the concepts and main ideas justifying image texture analysis in the spatial-frequency domain. Sections 3-6 introduce relevant properties of discrete wavelet, Gabor wavelet, dual-tree complex wavelet and contourlet transforms. Furthermore, numerous analytical methods and techniques used to extract texture features and perform similarity measurements are described. Section 7 discusses and briefly summarizes current achievements, while section 8 concludes the paper.

## 2. Texture-based Retrieval and Image Representation

Image texture can be seen as an image area containing repeated patterns of pixel intensities arranged in some structural way. Textures are prominent in natural images (as in grasslands, brick walls, fabrics, etc.) and many important properties for image description and interpretation are revealed through texture observation and analysis such as granularity, smoothness, coarseness, periodicity, geometric structure, orientation and so on. Therefore, texture retrieval is relevant to CBIR since texture characteristics are powerful in discriminating between images. As mentioned earlier, successful texture-based image retrieval relies strongly on effective feature extraction and similarity measurement steps yielding to good quality features that: a) contain *relevant texture information* and possess *high discriminating power* to ensure retrieval accuracy, b) can be structured in a *compact feature vector* with low dimensionality to shorten retrieval time, and c) possess *rotation invariance* property to recognize the rotated versions of the same texture sample.

Even though the concept of texture lacks formal and mathematical definition, there are renowned methodologies and approaches for texture feature extraction operating in the spatial domain (e.g. gray level co-occurrence matrices), in the frequency domain (e.g. Fourier spectrum measurements), or in the spatial-frequency domain (e.g. energy of wavelet coefficients) [4].

Spatial-frequency image transforms, also known as multiscale representations, decompose the image into a set of sub-images exhibiting image details and structures at multiple scales and multiple orientations. Each sub-image corresponds to a frequency sub-band; a localized partition of the image spectrum. Linear filter banks and down/up sampling operators are the main tools to perform such decompositions yielding various image representations with specific properties such as multiple scales, frequency selectivity, directional orientation, redundancy/compactness, perfect reconstruction, shift invariance and rotation invariance. Examples of spatial-frequency image transforms include discrete wavelets, steerable pyramids, Gabor wavelets, complex wavelets and contourlets.

Recent studies have reported the achievement of remarkable outcomes due to the development of a variety of new texture feature extraction techniques operating on these multiscale image representations. This probably was motivated by three main facts: a) the human visual system (HVS) adequacy to the spatial-frequency representation of signals, b) the inherent nature of texture patterns in terms of presence of edges, relation between primitive texture elements and variation in scales and orientations, and c) the psychological research on human texture perception which suggests that two textures are often difficult to discriminate when they produce a similar distribution of responses from a bank of linear filters [4, 5, 6].

Energy approach for texture feature extraction is very popular and consists in computing energy and characterizing its distribution through spatial-frequency sub-bands. A feature vector is formed from sub-band energy measures ($L^1$ norm, variance, standard deviation ...). Similarity measurement is commonly computed as the Euclidean distance, normalized Euclidean distance, or as the Manhattan distance between texture feature vectors of compared images.

Do and Vetterli [7] have introduced a generic statistical framework for texture retrieval considering, in a joint way, the formulation of the texture feature extraction and the similarity measurement problems. Starting from a statistical analysis of image data, a statistical model (SM) with good fits is suggested and texture feature vector components are derived from the estimated model parameters using the Maximum Likelihood estimator. The similarity measurement function is theoretically justified to have optimal properties if derived from the Kullback-Leibler (KL) divergences between the estimated model for the query and the estimated models for each image. The KL distance may be approximated by a Monte-Carlo optimization method if a closed form is not possible. This generic framework (SM-KL) has been very useful and gave rise to many renowned methods and techniques for texture retrieval in the space-frequency domain.

One can notice that the energy approach may be derived from the SM-KL statistical framework if Gaussian distribution (with mean, variance or standard deviation parameters) is assumed for each image sub-band.

The retrieval accuracy (or retrieval rate) is commonly measured as a *precision* parameter: the ratio between the number of retrieved relevant images and the total number of retrieved images. A recall parameter, calculated as the number of retrieved relevant images divided by the total number of relevant images, is also used to evaluate the retrieval efficiency.

## 3- Discrete Wavelets for Texture Retrieval

*3.1. The Discrete Wavelet Transform*

The two dimensional discrete wavelet transform (DWT) is an effective tool to analyze images in a multiscale framework and to capture localized image details in both space and frequency domains [8]. The DWT is efficiently implemented via the Mallat's tree algorithm that applies iterative linear filtering and critical down-sampling on the original image yielding three high-frequency directional sub-bands at each scale level in addition to one low-frequency sub-band usually known as image approximation. Directional sub-bands are sparse sub-images exhibiting image details according to horizontal, vertical and diagonal orientations. The decomposition process is illustrated in Figure 1 with the top image undergoing a first level decomposition to generate 3 detail sub-bands ($H_1$, $V_1$, and $D_1$), and one image approximation ($A_1$). At the second level of decomposition, the approximation image ($A_1$) undergoes the same process to produce a second scale level of image details ($V_2$, $H_2$ and $D_2$) and a new image approximation ($A_2$). The resulting 2-level DWT is shown in Figure 2.

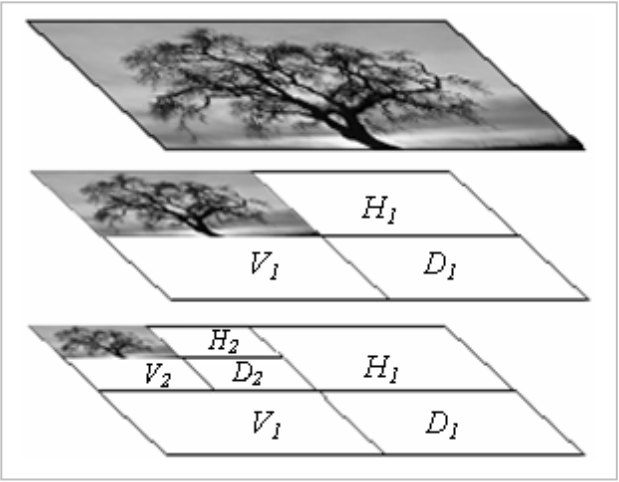

Figure 1: A two-level wavelet decomposition

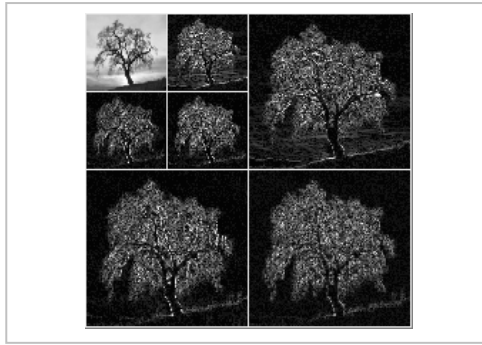
Figure 2: Resulting sub-bands for a 2-level DWT (original image in Figure 1)

In addition to being invertible and of perfect reconstruction, the DWT provides a highly compact image representation, that is, the transform is orthogonal, and incorporated down-sampling rates result into a total number of wavelet coefficients equal to the image size.

Since its development in the mid 80's, the DWT with its attractive properties has been very popular, extensively exploited in various fields of image processing and particularly successful in image compression applications. However, its use for texture analysis has revealed some limitations in capturing relevant information. In fact, major drawbacks are reported in many studies; lack of translation invariance (shift sensitivity), poor selectivity (only three detail sub-bands for each scale level) and poor directionality (only horizontal, vertical and diagonal orientations) [9, 10, 11]. In the last decade, many spatial-frequency transforms have emerged as improved extensions of the DWT.

*3.2. Variants of the DWT*

A wavelet packet transform (WPT) is a variant of the DWT which enhances frequency selectivity of wavelet coefficients. The implementation of WPT proceeds with a full decomposition of both details and approximation sub-bands of the original image instead of approximation decomposition alone [12]. Much larger number of decomposition sub-bands is provided at each scale level offering more specific details of any part of the frequency domain. The decomposition process diagram is illustrated in Figure 3 and Figure 4. When full decomposition is not required, various ways may be used to determine which sub-band is suitable for further decomposition.

The steerable pyramid [13] is another variant of the DWT using a set of orientation filters. Great directional-selectivity and approximate shift invariance are achieved inducing some cost relative to the substantial redundancy of wavelet coefficients. Notice that it is the choice of authors to consider the steerable pyramid as a variant of DWT. In other references, it may be presented as a new transform.

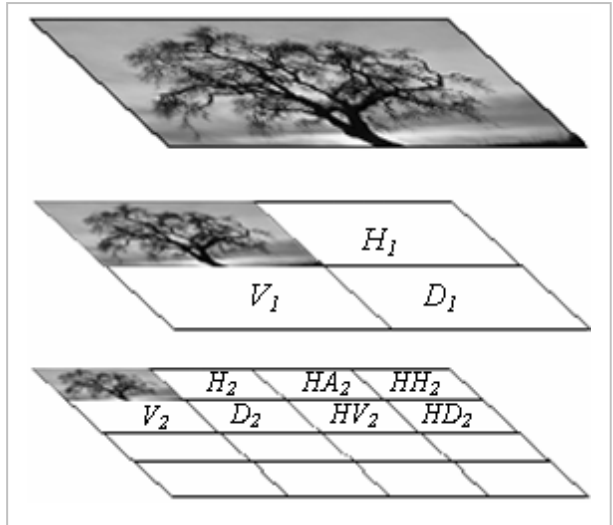
Figure 3: A two-level WPT decomposition

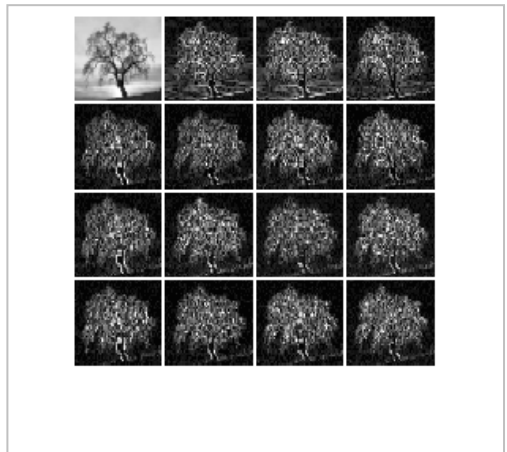
Figure 4: Resulting sub-bands of a 2-level WPT (original image in Figure 3)

*3.3. Wavelet-based Texture Retrieval*

A simple and popular approach for texture feature extraction in the wavelet domain is to form a feature vector from energy measures in wavelet sub-bands, using $L^1$ or $L^2$ norms, mean, variance, etc... Similarity measurement is commonly computed as the Euclidian distance (or normalized Euclidean distance) between compared feature vectors. The advantage of applying WPT for feature extraction resides in the flexibility to explore fine details from high and medium frequency subdivisions of the image instead of focusing on further decompositions of the lower image frequencies. This is important when knowing that textures may have relevant information in the middle and high frequencies [14]. The technique presented in [15] extracts texture features from WPT sub-band coefficients. First the image undergoes a preprocessing for texture denoising and enhancement. Then a 3-level WPT is applied to texture images producing 64 sub-bands. Finally, the mean and variance parameters are calculated for each sub-band to compose a texture feature vector of length 128. In [14], the mean of the magnitude of wavelet coefficients is calculated for each sub-band and added to the feature vector. It is shown that using a two-level WPT where further decompositions are applied to A, H and V sub-bands improves retrieval rates comparatively to a conventional DWT.
In [7], the statistical framework SM-KL is adopted and generalized Gaussian distribution (GGD) is found to be a good fit to each DWT sub-band. Model parameters are then reduced to the estimation of two values for each sub-band: the scale and the shape of the GGD. Under the assumption of sub-band independency, the similarity measurement is formulated as the overall KL divergence between all sub-band GGDs and result in a substantially simplified similarity metric. Moreover, considering particular cases of the GGD model (such as the Laplacian distribution) result in a KL-based similarity metric with closed form; that is involving the estimated model parameters (as feature vector components) in simple distance calculations.

Comparing the GGD-KL method to energy-based approaches on three-level wavelet decompositions, the computational load is quite comparable while retrieval rates on texture images increase significantly (from 65% to 77%) in favor of the GGD-KL method [7].

In [16], the statistical modeling of wavelet coefficients is enhanced by using a mixture of generalized Gaussian distribution. The resulting texture retrieval method MoGGD-KL outperforms a GGD-KL method in terms of retrieval rates (up to 20% of rate augmentation) at the cost of additional amount of calculations and increased size of feature vectors. In fact, the GGD-KL estimates 2 model parameters per wavelet sub-band while MoGGD-KL estimates $4K$ parameters per wavelet sub-band where $K$ is the number of fitting GGDs in the mixture. The value $K=2$ is often used, thus resulting in feature vectors of length $24L$, with $L$ being the number of wavelet decomposition levels.

In [17], the application of the GGD-KL method, previously described, is extended to colored texture images. Since RGB color components are highly correlated, each RGB image is first decorrelated into three new independent layers using an independent component analysis (ICA) algorithm. Afterwards, Haar wavelet decomposition is applied on each layer and each resulting sub-band is modeled using the GGD-KL method. Two parameters are estimated for each sub-band GGD and added to the feature vector. Finally the database images with the smallest KL divergence to the query image are retrieved. Simulation results show the advantage of using the independent component color space over its RGB counterpart.

In [18], the GGD-KL method is applied to the wavelet sub-bands of a steerable pyramid [13]. Since the steerable pyramid is an over complete transform with a redundant content, the assumption of sub-band independency does not hold. In order to ensure accurate statistical modeling, the inter-band correlation is reduced by applying independent component analysis (ICA) on the wavelet coefficients. Afterwards, the GGD-KL method is employed to model the resulting data. For each image, the feature vector is composed of its estimated GGD parameters in addition to the ICA filtering matrix. The whole framework is illustrated in Figure 5. Experiments on sets of texture images decomposed into steerable pyramids with two scales and four orientations demonstrated better results for the GGD-KL-ICA over methods using simply GGD-KL.

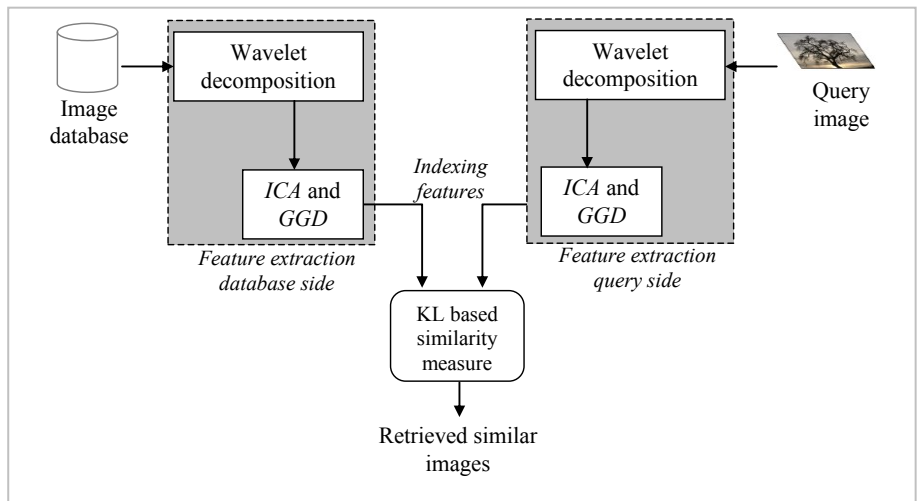

Figure 5: Texture feature retrieval diagram using ICA and GGD

In [19], Do and Vetterli enhance their statistical modeling by using a wavelet domain hidden Markov model (WD-HMM) that captures sub-band marginal distributions as well as strong inter-scale and inter-orientation dependencies of wavelet coefficients. Replacing the DWT by a steerable pyramid transform leads to rotation-invariant texture image retrieval where image feature vector is represented as a covariance matrix containing WD-HMM model parameters. While estimating model parameters may require substantial computational load for algorithm training, the KL-similarity measurement is approximated with a fast algorithm based on a Monte Carlo method that is, comparable to the computation of the Euclidean distance between two feature vectors.

Another alternative for statistical modeling has been recently presented in [20]. A multivariate sub-Gaussian model is tied up to the sub-band coefficients of a steerable pyramid providing a statistical description of the non-Gaussian nature of coefficient distribution in addition to the interactions and inter-dependencies across scales and orientations. For each image, the derived texture feature vector is a set of covariation matrices. Thus, the similarity function is formulated as a matrix-based norm which achieves rotation invariance by means of angular alignment of the two feature vectors being compared. Based on achieved retrieval performance, the proposed method is comparable to the WD-HMM method in [19] while maintaining reasonably a lower computational complexity.

Since the wavelet transform and its variants (wavelet packets, steerable pyramid) have existed for over a decade, the research involved with this transform have had sufficient time to mature into powerful statistical modeling frameworks and highly effective techniques for image texture retrieval. Moreover, a thorough understanding of its strengths and weaknesses as well as their impact on image analysis and feature extraction has contributed to the emergence of new spatial-frequency transforms with desired properties.

## 4. Gabor Wavelets for Texture Features Analysis

*4.1. Gabor Wavelet Transform*

Gabor wavelet transform (GWT) is a classic method for multichannel, multiresolution analysis that represents image variations at different scales. Gabor filters, as shown in Figure 6, are a group of wavelets obtained from the appropriate dilation and rotation of Gabor function: a Gaussian modulated sinusoid. By capturing image details at specific scales and specific orientations, Gabor filters present a good similarity with the receptive fields on the cells in the primary visual cortex of the human brain. Gabor wavelet transform provides a flexible method for designing efficient algorithms to capture more orientation and scale information [21].

Many researches stated that Gabor wavelet transform represents one of the efficient techniques for image texture retrieval yielding good results in content-based image retrieval applications due to many reasons [21, 22, 23]:
- Being well suited for image signal expression and representation in both space and frequency domains.
- Presenting high similarity with human visual system as stated above.
- Offering the capacity for edge and straight line detection with variable orientations and scales.
- Not being sensitive to lighting conditions of the image.

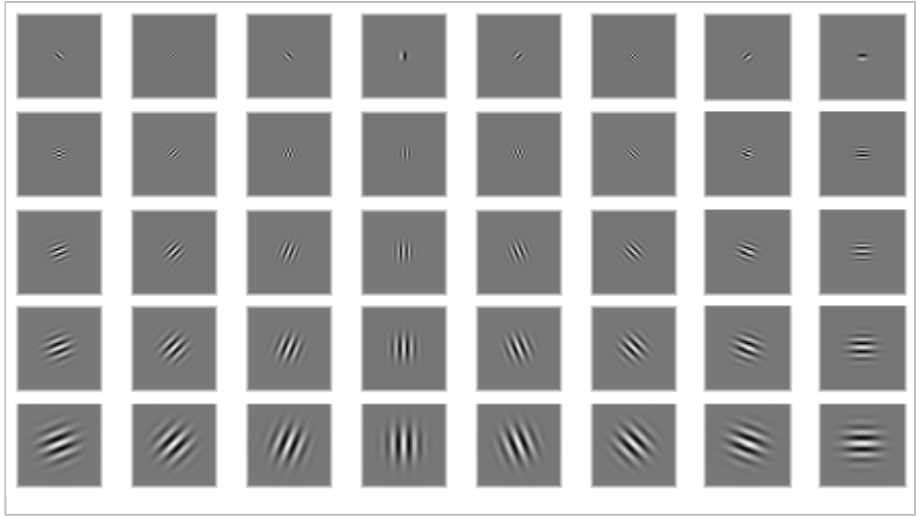

Figure 6: A set of real impulse responses: multiscale, multi-orientation Gabor wavelet filters

On the opposite side, Gabor wavelet representations suffer from some weaknesses such as their costly computational complexity, their non invariance to rotation as well as the non orthogonal property of the Gabor filters that implies redundancy in the filtered images. As each Gabor filter or wavelet captures a specific frequency and a specific direction from the image, texture feature extraction using this representation requires the re-convolution of the image with the Gabor function for each change of one of these parameters.

A simplified Gabor wavelet transform (SGWT) has been introduced to enhance efficiency and reduce computation complexity inherent to the use of multiscale multi-orientation Gabor filters for texture feature extraction [24].

To generate a simplified Gabor wavelet (SGW), the continuous values of the Gabor wavelet are quantized to a selected number of levels: 4 levels in the example shown in Figures 7 and 8. This generates a set of ellipses representing the different quantization levels of the positive and negative values of the SGW as shown in Figure 7 and Figure 8. The chosen quantization level impacts on the number of rectangles obtained in the SGW. The higher this number is, the more similar is the SGW with the original GW with all the proportional associated computation complexity.

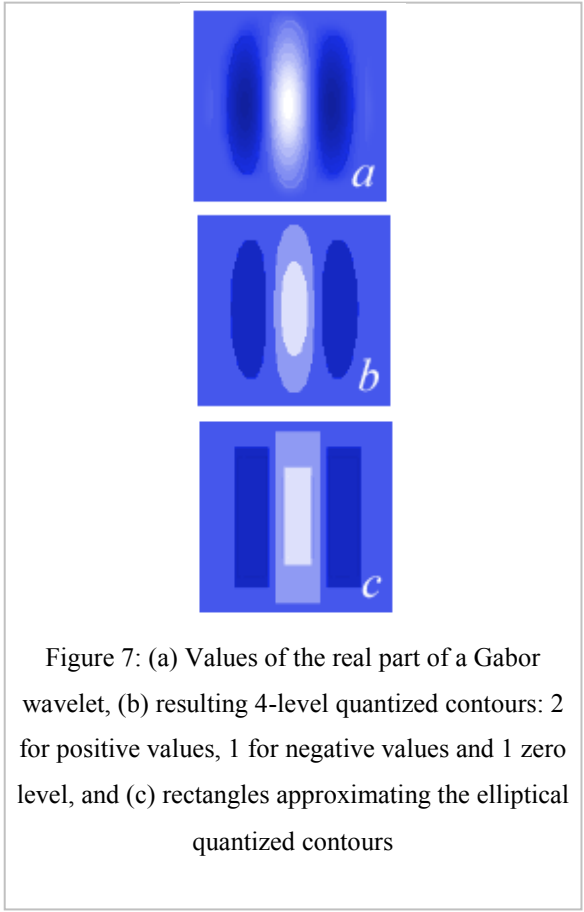

Figure 7: (a) Values of the real part of a Gabor wavelet, (b) resulting 4-level quantized contours: 2 for positive values, 1 for negative values and 1 zero level, and (c) rectangles approximating the elliptical quantized contours

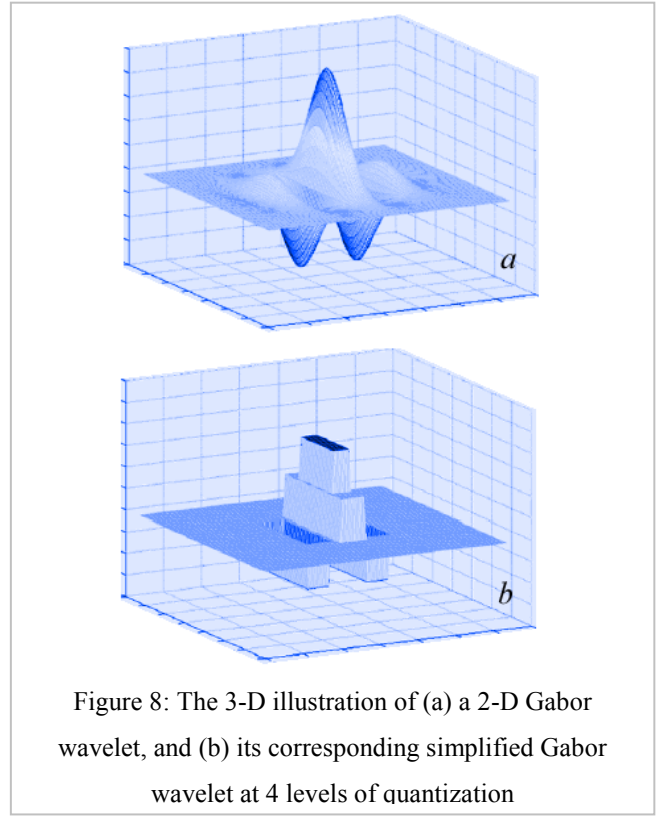

Figure 8: The 3-D illustration of (a) a 2-D Gabor wavelet, and (b) its corresponding simplified Gabor wavelet at 4 levels of quantization

*2.2. GWT Texture Features*

Most proposed texture retrieval methods on GWT are based on the energy approach [23, 25, 26, 27, 28, and 29]. Texture extraction yielding rotation invariant features has been studied in the context of Gabor wavelets to cope with the problem of non correspondence between similarity measurements of texture retrieval from the same image undergoing different rotations. For example, images in Figure 9 *(a)* and 9 *(c)* are the same with different orientations but will have very large distance if the texture similarity measurement is applied directly. The proposed solution in [30, 31] is a normalization approach to solve the rotation variant problem associated with Gabor texture features. The total energy for each orientation is calculated, the dominant orientation is designated based on the highest total energy, and the feature vector is circularly shifted to get the feature elements of the dominant orientation moved to the first positions. The other elements are circularly shifted accordingly [31]. Experiments conducted on natural and texture images yield interesting results with 96% accuracy on a database of 1000 texture images. However, more detailed studies have to be conducted based on natural images with more details regarding achieved improvements related to computation, time wise and complexity wise.

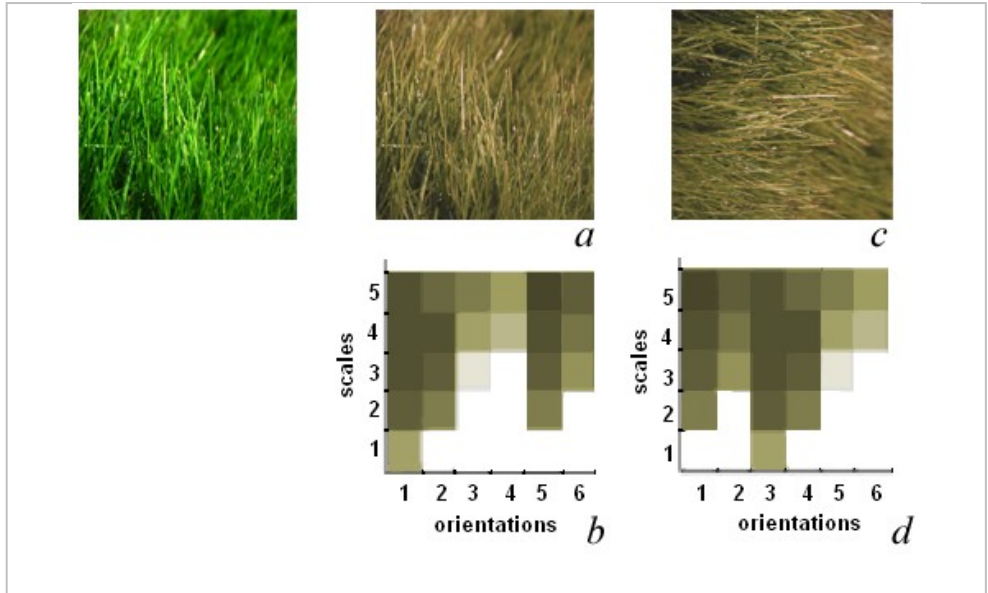

Figure 9: Two grayscale texture images (*a* and *c*) with their respective feature maps (*b* and *d*), the second image is a 90° rotation of the first image. (Original image on top left)

In [32], retrieval performance and rotation invariance are substantially improved when standard similarity metrics are replaced by a multivariate statistical Wald-Wolfowitz test: a non-parametric distance based on the concept of minimum spanning tree.

In their paper, Sastry et al. [33] presented a texture extraction method which takes advantage of the Gabor filters scalability without being limited by their rotation dependency. A proposed modified Gabor wavelet transform (MGWT) is therefore set in such a way that it becomes rotation independent. This is done by taking integration of the Gabor function angle over all possible angles keeping the complex exponential frequency constant, thus, allowing texture feature analysis for a single frequency and multiple scales. In contrast with standard Gabor wavelet transform and according to [33], this technique can achieve rotation invariance of extracted features even though it requires computational overheads. More research contributions on Gabor wavelets for extracting texture features with both rotation and scale invariance can be found in [25, 26].

As mentioned earlier, using simplified Gabor wavelet transform instead of GWT helps reduce the computation complexity inherent to the multiscale multi-orientation Gabor filters for texture feature extraction. However, Choi et al. [24] state that the SGWT presents a drawback as it is sensitive to lighting conditions in the image. This is due to the fact that the formed SGW has a non-zero mean. To cope with this, the SGW has to be demeaned by subtracting the coefficients of the SGW from its mean value. Experimental results showed that, with the same number of scales and orientations, the use of SGW yielded results similar or slightly better than the ones obtained using Gabor Wavelet, with the exception for the images with high variations in lighting for which the GWT outperforms the SGWT.

Despite its costly computational complexity, Gabor wavelet texture retrieval methods covered above as well as other studies have proven efficiency for texture retrieval and have presented interesting capabilities for rotation invariance. It is also worthy to mention that, in general, using a GWT for texture retrieval helps cope with the limits of DWT with respect to its poor frequency selectivity and orientation.

## 5. Complex Wavelets for Texture Retrieval

### 5.1. The Dual-Tree Complex Wavelet Transform

The dual-tree complex wavelet transform (DT-CWT) as introduced by Kingsbury [34] has been found to be a useful tool for image texture analysis and feature extraction. The DT-CWT can be seen as an image transform which assembles many attractive properties of both Gabor and discrete wavelet transforms while avoiding many of their drawbacks.

The DT-CWT pyramid structure is much faster to implement than a Gabor wavelet transform while having a Gabor-like frequency response with six multiscale directional sub-bands at (±15°, ±45°, ±75°) as shown in Figure 10. Indeed, the DT-CWT is efficiently implemented as four parallel real DWT filtering trees of linear complexity and critical sampling, resulting in complex valued coefficients. At the cost of a reasonable amount of additional computational load and a 4:1 redundancy, the DT-CWT provides two main advantages over the real DWT image representation; namely, approximate shift invariance and improved directional selectivity which renders DT-CWT nearly rotation invariant [34]. All these good properties are known to be particularly suitable for texture feature extraction and representation. This is the reason that motivated many researchers in developing multiscale texture analysis and retrieval techniques on the DT-CWT. Moreover, many variants of this transform are introduced in order to improve retrieval efficiency. In [35], a double dyadic DT-CWT ($DT^3$-CWT) is built from the DT-CWT to allow a fine signal analysis at and between dyadic scales. In [10], directional selectivity is extended to twelve directions, namely (0°, ±15°, ±30°, ±45°, ±75°, 60°, 90°, 120°). This is achieved by designing rotated complex wavelet filters (RCWF) and adding a DT-RCWF pyramid as a complement to the existing DT-CWT pyramid.

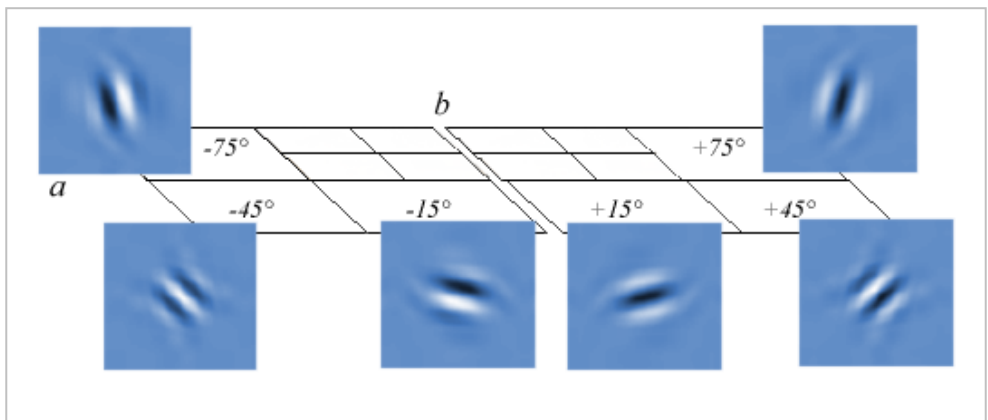

Figure 10: 2D DT-CWT filters; a) Real impulse responses (space domain), b) Ideal support of frequency responses. Six directional orientations are provided (±15°, ±45°, ±75°)

### 5.2. Extracting Texture Features from a DT-CWT

Many proposed techniques illustrate the use of the DT-CWT in extracting texture features and improving feature discrimination and retrieval. So far, computing energy and characterizing its distribution through the multiscale directional sub-bands is most probably the dominant approach for texture feature extraction in the DT-CWT domain. In [36], the most significant 16 DT-CWT sub-bands in terms of energy are selected for feature extraction. The mean and standard deviation of each sub-band are calculated and arranged in a feature vector of length 32. Experimental results show the capacity of such features in discriminating among various texture patterns. A quite similar technique is proposed in [37] where each feature vector is derived from all available detail sub-bands. A Canberra distance is employed for similarity measurement. It is shown that the DT-CWT outperforms real DWT in terms of retrieval performance for Brodatz texture samples but no results are provided on natural images. Kokare et al. [10] exploit jointly a DT-CWT and a set of dual-tree rotated complex wavelet filters (DT-RCWF) to provide a multiscale image representation with greater directional selectivity. Then, robust rotation invariant texture features are accurately extracted from each scale level by taking average energy and average standard deviation of twelve oriented sub-bands having each a distinct direction. Texture features are also obtained from the approximation sub-band at the lowest scale level. Experimental results on rotated and real world textured images show that the proposed method improves retrieval rates and outperforms a traditional DWT-based approach. Lo et al. [35] derive scale and rotation invariant texture features from the energy of a double dyadic DT-CWT (DT$^3$-CWT).

In [38], the authors propose a specific implementation of the statistical framework SM-KL for texture retrieval in the DT-CWT domain. The marginal density of the coefficient magnitudes of the detail sub-bands are shown to be efficiently modeled using either a two parameter Weibull distribution or a two-parameter Gamma distribution. Assuming sub-band independency, a similarity measure between two images is computed as the overall Kullback-Leibler (KL) divergence between all sub-bands. Even if sub-band independency assumption is questionable, comparison of retrieval performances shows the superiority of the Weibull model while Gamma Model is still satisfactory. Moreover, it is shown that using a classic approach with mean and standard deviation features from the DT-CWT is computationally more appealing and lead to better retrieval results than the combination of a DWT with a generalized Gaussian distribution. Significant improvements in retrieval rates are achieved in a recent work of the same authors [39] while extending the processing to RGB color images. Indeed, sub-band dependency property is taken into account and a multivariate modeling by means of Copulas is found to be a good fit to DT-CWT coefficient magnitudes. For similarity measurement, a Monte Carlo approach is employed to approximate the KL divergence between two Copula models. Even if image color is a very important attribute for content retrieval, the paper [40] analyses color-based texture features on the DT-CWT and highly recommends separate rather than joint feature processing for color and texture information.

Classic feature vectors derived from energy measurements in the DT-CWT domain are shown to have good capacity in discriminating among various texture patterns and lead to better retrieval results than energy approach on DWT. Despite of the many attractive attributes offered by the DT-CWT, its usage for texture analysis is still in its infancy and lacks more research and investigation to achieve the maturity known for other transforms.

## 6. Contourlets for Texture Retrieval

### 6.1. The Contourlet Transform

The contourlet transform (CT) is a 2D directional multiscale image decomposition which has been introduced to overcome the DWT inefficiency in terms of directional selectivity [11]. It is constructed by combining two distinct and successive decomposition stages: a multiscale decomposition followed by a directional decomposition. First, a multiscale decomposition uses a Laplacian pyramid scheme to transform the image into one coarse version plus a set of Laplacian sub-images (LP). Second, a directional stage applies iteratively two dimensional filtering and critical down sampling to further partition each LP sub-band into different and flexible number of frequency wedge-shaped sub-bands, thus capturing geometric structures and directional information in real-world images (see Figure 11). In addition to offer perfect reconstruction and high computational efficiency, the CT is almost critically sampled with a small redundancy factor of up to 4:3 due to the Laplacian pyramid. When compared to the DWT, the CT with its extra feature of directionality yields some improvements and new potentials in image analysis applications. Indeed, various experiments clearly show that smooth object boundaries are efficiently represented by a small number of local coefficients in the right directional sub-bands, leading to better reconstruction of fine contours [11]. The above-mentioned statements provide enough motivation to use contourlets in extracting significant texture image features and to determine how effective it is.

In general, hierarchical processing which requires cooperation or mapping between various image scale levels is facilitated and rendered more accurate when all these levels are of the same size. This is the reason why a redundant contourlet transform (RCT) and a non subsampled contourlet transform (NSCT) variants have been introduced providing equal-size directional sub-band images [42, 43]. In the RCT variant, redundancy is achieved by discarding down-sampling operations at the Laplacian stage while in the NSCT decomposition; all down-sampling operations are discarded thus reducing significantly aliasing problems and allowing for fully shift- invariance.

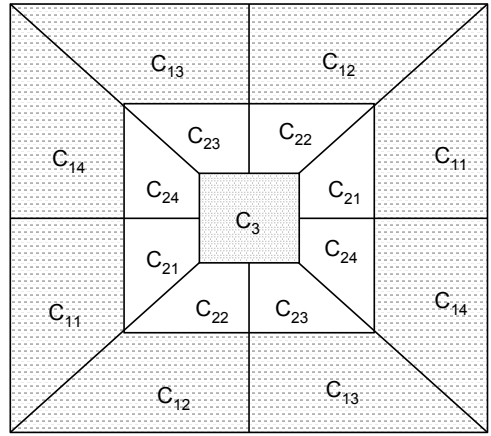

Figure 11: Frequency partition of a contourlet decomposition with 3 scale levels and 4 frequency directions. The coarse image approximation ($C_3$) is not decomposed directionally

*6.2. Extracting Texture Features from Contourlet Image Representation*

In the context of the statistical framework SM-KL, Po and Do [43] have conducted qualitative and quantitative studies on the statistical modeling of contourlet coefficients in natural images. New properties are revealed such as: a) contourlet coefficients strongly depend on their spatial, multiscale and directional neighborhood, especially for highly textured images, and b) conditioned on their neighborhood, contourlet coefficients are approximately Gaussian. Based on these new properties, a hidden Markov tree (HMT) model with Gaussian mixtures is tied up to each CT sub-band. Estimated by an iterative expectation maximization algorithm, model parameters for all sub-bands are adopted as a texture feature vector (three parameters from the coarse sub-band and four from each detail sub-band). For similarity measurement between two feature vectors, a Kullback-Liebler distance is suggested. The contourlet HMT is shown to be more efficient than wavelet HMT in terms of retrieval rates when searching textured images with diverse high directionalities (other than vertical and horizontal ones). However, rotation invariance capability is not addressed. This method has been adapted to intra-band dependencies of luminance images [44]. A Gaussian mixture model is suggested for each CT sub-band and its parameters (means, variances/covariances) are adopted as a feature vector. Model parameters are estimated using the Markov dependencies of neighboring coefficients according to the directional orientation in each sub-band. Derived feature vectors may contain 72 elements for a 3-level CT having 40 sub-bands. Further extensions exploiting intra-band dependencies in chrominance components allow better retrieval rates at the cost of larger feature vectors (around 240 elements per vector) [44].
In [45, 46], the marginal distribution of contourlet coefficients is modeled as a generalized Gaussian density (GGD) in each sub-band and KL distance is taken as a similarity measure. Two parameters in this model, namely the scale and the shape, are estimated for each sub-band and adopted as part of the texture feature vector [46]. Energy-based approach is applied successfully on CT sub-bands [47, 48]. Assuming a Gaussian distribution for each contourlet sub-band, the texture feature vector in [47] is formed as a concatenation of the mean, standard deviation, maximum value and minimum value of all sub-bands. A simple Manhattan distance ($L^1$ norm) between two feature vectors is considered for similarity measurement. In [48], a feature vector is created as a normalized set of all sub-band standard deviations. Retrieval performance is further improved when using the Manhattan distance instead of the Euclidean distance for similarity measurement.

The proposed method in [49] takes advantage of CT directional selectivity in each contourlet scale level to reduce the rotation variance of extracted features. Hence, at each scale level, standard deviation σ is estimated for each sub-band, then the maximum value of all σ and the sum of all σ are selected to create a feature vector of size *2M* (for a *M* level decomposition). A Manhattan distance between feature vectors is employed to find similar images. Simulation results on rotated Brodatz texture images show that increasing the number of directions in each scale level improves the retrieval accuracy. This method is particularly attractive when considering its low computational load and the compactness of derived texture feature vectors.

The contourlet transform as a compact multiscale image representation with high degree of directionality is relevant to texture analysis and feature extraction in natural images. Yet, existing studies have shown its superiority to other image transforms. Based on the first results achieved so far, the future of texture feature extraction in the contourlet domain is very promising and deserves much more research efforts to achieve one of the most significant aims, namely the efficient content-based image retrieval in large-scale image databases.

## 7. Discussion

Spatial-frequency image representations play a critical role for texture feature extraction and characterization. Table 1 summarizes relevant properties such as transform redundancy and directional selectivity which have great impact on retrieval efficiency and the achievement of rotation invariance. Two broad approaches for texture retrieval in the spatial-frequency domain are generally adopted: energy-based approach and statistical modeling approach mainly based on the SM-KL generic framework.

The energy-based approach assumes that different texture patterns have different energy distribution in the space-frequency domain. This approach is very appealing due to its low computational complexity involving mainly the calculation of first and second order moments of transform coefficients. Widely applied on most known spatial-frequency transforms, it achieved reasonable performance in terms of retrieval rates as well as computational load. Despite the fact that its associated formalism does not explicitly take into account the rotation invariance constraint, this is still achievable when using adequately spatial-frequency transforms with high degree of directionality such as in the GWT [23], DT-CWT [10] and in contourlets [49].

However, the statistical modeling approach leads to more complete and more rigorous theoretical frameworks that a) handle both feature extraction and similarity measurement problems in a joint way, b) capture and integrate the specific characteristics of the spatial-frequency transform, and c) incorporate explicitly the rotation invariance constraint of texture retrieval. Consequently, most derived techniques are theoretically justified and validated but unfortunately, their computational cost, in general, is much more expensive than energy-based ones.

Table 2 is a brief summary of the most recently introduced statistical models for texture retrieval in the spatial-frequency domain, as presented in previous sections, as well as an attempt to quantify the dimensionality of extracted feature vectors. One can notice that only few statistical methods are proposed for DT-CWT and CT in comparison with the DWT. Moreover, no method that uses CT or DT-CWT achieves rotation invariance. Despite the attractive properties of these new transforms and their important potential impact on texture retrieval, they are still in their early age of research and development.

TABLE 1 – Comparison of relevant properties of spatial-frequency transforms

| Image Transform | Transform Variants | Transform Redundancy | Directional Selectivity | Shift or Rotation Invariance |
|---|---|---|---|---|
| Discrete Wavelet Transform (DWT) [8] | | No | Poor (3 orientations) | No |
| | Wavelet Packet Transform (WPT), [12] | Yes (flexible) | Medium (flexible) | No |
| | Steerable Pyramid [13] | Yes (flexible) | Rich (flexible) | Yes |
| Gabor Wavelet Transform (GWT) | | Yes | Rich (flexible) | No |
| | Simplified Gabor Wavelet Transform (SGWT), [24] | Yes | Rich (flexible) | No |
| | Modified Gabor Wavelet Transform (MGWT), [33] | Yes | Rich (flexible) | No |
| Dual Tree-Complex Wavelet Transform (DT-CWT) [34] | | Yes (4:1) | Medium (6 orientations) | Yes |
| | Dual Tree-Rotated Complex Wavelet Filters (DT-RCWF) [10] | Yes (8:1) | Rich (12 orientations) | Yes |
| | Double Dyadic Dual Tree-Complex Wavelet Transform (DT$^3$-CWT), [35] | Yes (8:1) | Medium (6 orientations) | Yes (approximate) |
| Contourlet Transform (CT) [11] | | Yes (4:3) | Rich (flexible) | No |
| | Redundant Contourlet Transform (RCT), [42] | Yes (more than 4:3) | Rich (flexible) | No |
| | Non Sub-sampled Contourlet Transform (NSCT), [43] | Yes (more than 4:3) | Rich (flexible) | Yes |

TABLE 2 – Statistical models for texture retrieval in the spatial-frequency domain. Energy approaches are not included.

| Image Transform | Model/Method | Feature Vector | Rotation Invariance | References |
|---|---|---|---|---|
| DWT ($L$ scale levels) | GGD-KL | scale+shape (2/sub-band) | No | [7] |
| | GGD-KL-ICA (RGB) | scale+shape (2/sub-band) | No | [17] |
| | GGD-KL-ICA (steerable) | scale+shape (2/sub-band) + ICA matrix | No | [18] |
| | WD-HMM-KL (DWT and steerable) | covariance matrix ($14L-1$ param. for $L$ scales) | Yes | [19] |
| | MoGGD-KL | $4K$/sub-band (mixtures of $K$ GGDs/sub-band) | No | [16] |
| | Multivariate sub-Gaussian (steerable) | $L$ covariation matrices $JxJ$ (for $J$ orientations and $L$ scales) | Yes | [20] |
| GWT | Multivariate Wald-Wolfowitz test | mean+variance (2/sub-band) | Yes | [32] |
| DT-CWT ($L$ scale levels) | Weibull-KL | scale+shape (2/sub-band) | No | [38] |
| | Gamma-KL | scale+shape (2/sub-band) | No | [38] |
| | Copula-KL (RGB) | 2 Weibull param./sub-band + definite matrix | No | [39] |
| CT ($L$ scale levels) | GGD-KL | scale+shape (2/sub-band) | | [45], [46] |
| | MoG-HMT-KL (inter-band) | (3/coarse band + 4/detail sub-band) | No | [43] |
| | MoG-HMT-KL (intra-band) | ($L=3$, 72 elements) (240 elements for RGB) | No | [44] |

- GGD : Generalized Gaussian Distribution
- MoGGD: Mixture of Generalized Gaussian Distributions
- KL: Kullbback-Leibler distance (or divergence)
- ICA: Independent Component Analysis
- WD-HMM: Wavelet Domain Hidden Markov Model
- HMT: Hidden Markov Tree
- MoG-HMT: Mixture of Gaussian –Hidden Markov Tree
- GWT : Gabor Wavelet Transform
- DWT: Discrete Wavelet Transform
- DT-CWT: Dual Tree-Complex Wavelet Transform
- CT: Contourlet Transform

## 8. Conclusion

In this paper we presented a brief overview of some of the latest methodologies and techniques relevant to texture-based image retrieval using spatial-frequency image representations. We tried to provide a general picture of the trends in this emerging field of research further develop a vision for future in-depth explorations and suggest design choices for texture retrieval applications. As mentioned before, CBIR systems require high retrieval efficiency while optimizing time performance and preserving low computational complexity. In general, spatial-frequency image decomposition is useful for texture feature extraction and discrimination at the cost of additional amount of computation resources. New texture retrieval techniques try to exploit more sophisticated spatial-frequency transforms that provide a great directional selectivity such as dual-tree complex wavelets and contourlets. Although scale and directionality have a positive impact on the quality of texture feature extraction and help achieving rotation invariance, it is still early to talk about a comparative performance study of these methods due to the absence of standard test-beds and uniform evaluation criteria that can operate evaluations based on the performance factors mentioned earlier.